\documentclass[conference]{IEEEtran}
\IEEEoverridecommandlockouts
\usepackage{cite}
\usepackage{amsmath,amssymb,amsfonts}
\usepackage{algorithmic}
\usepackage{graphicx}
\usepackage{textcomp}
\usepackage{xcolor}
\usepackage{subcaption}
\usepackage{amsmath,amssymb}
\usepackage{amsthm}
\usepackage{algorithm}
\usepackage{dsfont}

\usepackage[colorinlistoftodos]{todonotes}

\def\BibTeX{{\rm B\kern-.05em{\sc i\kern-.025em b}\kern-.08em
    T\kern-.1667em\lower.7ex\hbox{E}\kern-.125emX}}
\begin{document}

\title{Pedagogical Demonstrations and Pragmatic Learning in Artificial Tutor-Learner Interactions}

\author{\IEEEauthorblockN{Hugo Caselles-Dupré, Mohamed Chetouani, Olivier Sigaud}
\IEEEauthorblockA{Sorbonne Université, CNRS, Institut des Systèmes Intelligents et de Robotique (ISIR)}
}

\maketitle


When demonstrating a task, human tutors pedagogically modify their behavior by either ``showing" the task rather than just ``doing" it (exaggerating on relevant parts of the demonstration) or by giving demonstrations that best disambiguate the communicated goal \cite{ho2016showing, shafto2014rational}. Analogously, human learners pragmatically infer the communicative intent of the tutor: they interpret what the tutor is trying to teach them and deduce relevant information for learning\cite{gweon2021inferential, gweon2010infants, navarro2012sampling}. Without such mechanisms, traditional Learning from Demonstration (LfD) algorithms will consider such demonstrations as sub-optimal \cite{ravichandar2020_lfd}. In this paper, we investigate the implementation of such mechanisms in a tutor-learner setup where both participants are artificial agents in an environment with multiple goals. Using pedagogy from the tutor and pragmatism from the learner, we show substantial improvements over standard learning from demonstrations.

More specifically, we make the following contributions:
(i) we adapt pedagogical tutors and pragmatic learners ideas from developmental psychology to AI in a collaborative training setup~; (ii) we introduce pedagogy teaching and pragmatic learning mechanisms that are general enough to be applied to any artificial agent policy learning scenario (multi-arm bandit, evolutionary strategies, reinforcement learning)~; (iii) we provide an analysis of the benefit of adopting such approaches in the context of teachable and autotelic agents.

\section{Experimental tutor-learner setup}

\begin{figure}
    \centering
    \includegraphics[scale=0.4]{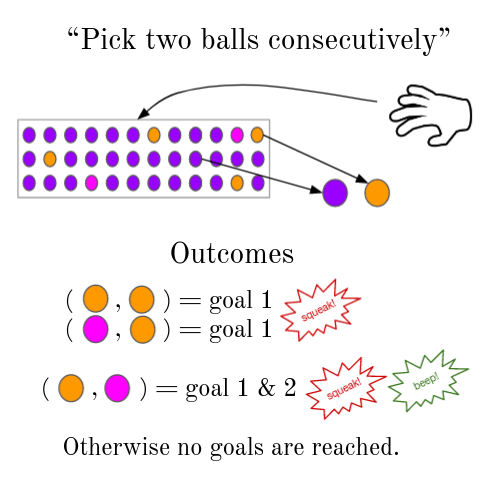}
    \caption{Our "Draw two balls" environment. An agent draws two balls from the box consecutively, and obtains a squeaking sound if goal 1 is achieved, a beeping sound if goal 2 is achieved, and nothing otherwise. Notice that picking the orange ball and then the pink ball reaches both goal 1 and goal 2, a source of goal ambiguity for demonstrators.}
    \label{fig:ballsenv}
\end{figure}

Our collaborative tutor-learner setup is depicted in \figurename~\ref{fig:ballsenv}. It is inspired from the squeaking balls environment presented in \cite{gweon2021inferential}. In our variant, there is a bucket of purple, orange and pink balls to choose from. Purple balls are more numerous than any other balls, an information that the agents have. The agents must pick two balls consecutively, and upon their choice, they can obtain three possible outcomes. If the picked balls are (orange, orange) or (pink, orange), goal 1 is reached. If the picked balls are (orange, pink), goal 1 and goal 2 are achieved. Otherwise, no goal is reached and nothing happens. The achievement of the goals are communicated to the agent.


\subsection{Policy learning}
\label{sec:policy}

The tutor and the learner are both represented by their respective goal-conditioned policies $\pi_T(.|g_t)$ and $\pi_L(.|g_l)$. These policies might be learned using multi-armed bandits, reinforcement learning or evolutionary strategies. The tutor's policy is already learned when it provides demonstrations to the learner.

In our experiments, a policy is a probability distribution for the first ball pick (purple, pink or orange), plus three policies conditioned on the first pick provided the probability distribution of picking the second ball given the first. There are three possible outcomes: no goals achieved, goal 1 and goal 2.

\subsection{Learning from demonstrations: training loop}

At each iteration, the tutor chooses a desired goal $g_d$ it wants the learner to achieve it on its own. It then presents a demonstration $D(g_d)$ for the chosen goal to the learner, which makes a prediction $\widehat{g_d}$ of the associated goal based on the demonstration. Then the learner plays and explores the environment using its policy conditioned on the predicted goal $\pi_L (.|\widehat{g_d})$ and obtains a trajectory $traj_L(\widehat{g_d})$ resulting in an achieved goal $g_a$. Finally, the learner is rewarded if $g_d = \widehat{g_d}$ and if $\widehat{g_d} = g_a$. It updates its policy using these two feedback signals.

Such autotelic (capable of generating and pursuing its own goals) and teachable (benefits from teaching via natural social interactions) agents present abilities that previous research argue are key steps to endow agents with the necessary capabilities to reach general intelligence \cite{sigaud2021towards, akakzia2021grounding, colas2020language, akakzia2022help}.

\section{Adding pedagogy and pragmatism}

We now present how the tutor can add pedagogy to its teaching signals to better convey important information about the goals, and how the learner can pragmatically interpret this implicit information in the teaching signals to learn better.

\subsection{Teaching with pedagogy}

Psychological research \cite{ho2016showing} has shown that human tutors often emphasize and exaggerate certain crucial aspects of the task to better demonstrate it, not with words but with actions. The key insight is that helpful tutors do not merely select optimal demonstrations of a task, but rather choose demonstrations that best disambiguate a goal from other goals. 

For the tutor, goal disambiguation results from using a goal prediction reward. When the pedagogical tutor learns to solve the task, it modifies its policies to better achieve the goal, like a naive tutor, but also when it produces a trajectory with which it can predict the pursued goal. This way, the learner will better predict the right goal using the pedagogical demonstration.

\subsection{Learning with pragmatism}

If the tutor demonstrations are biased towards actions that are unlikely for a naive prior, the learner can detect this pedagogical intention and deduce relevant information from it \cite{gweon2021inferential}. In our case, the purple balls being more frequent, if the tutor does not pick them often, the learner can detect this strong sampling bias and deduce that picking the purple balls is not relevant for goal achievement. 

We implement this using a prior mean probability of demonstration $D_{mean}$ using the probability of picking each colored ball. If the presented demonstration probabilities are inferior to $D_{mean}$ five times in a row, then strong sampling bias is detected and the pragmatic learner increases its probability of picking the purple ball for achieving no goal.


\section{Results}

\begin{figure}

  \begin{subfigure}[b]{1\columnwidth}
    \includegraphics[width=\linewidth]{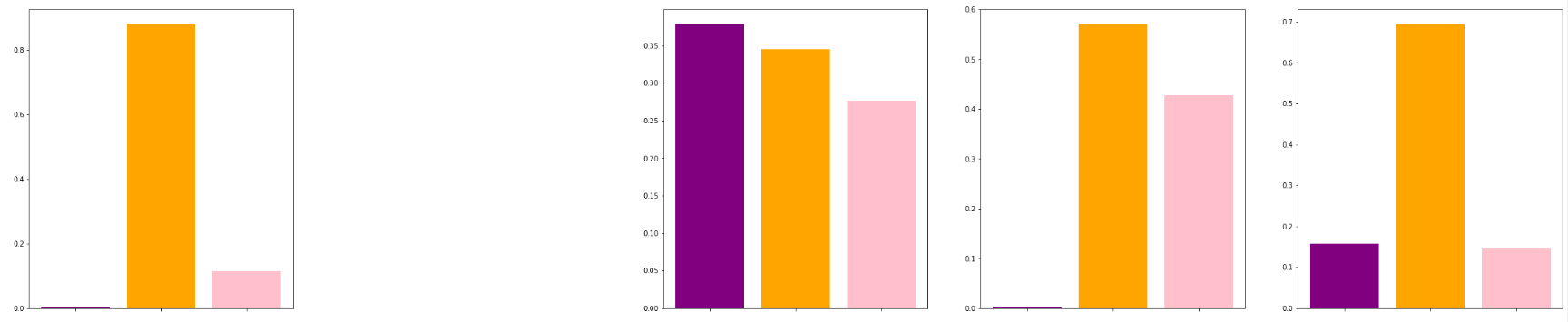}
    \caption{Naive tutor policy for goal 1.}
  \end{subfigure}
  \hfill 
  \begin{subfigure}[b]{1.\columnwidth}
    \includegraphics[width=\linewidth]{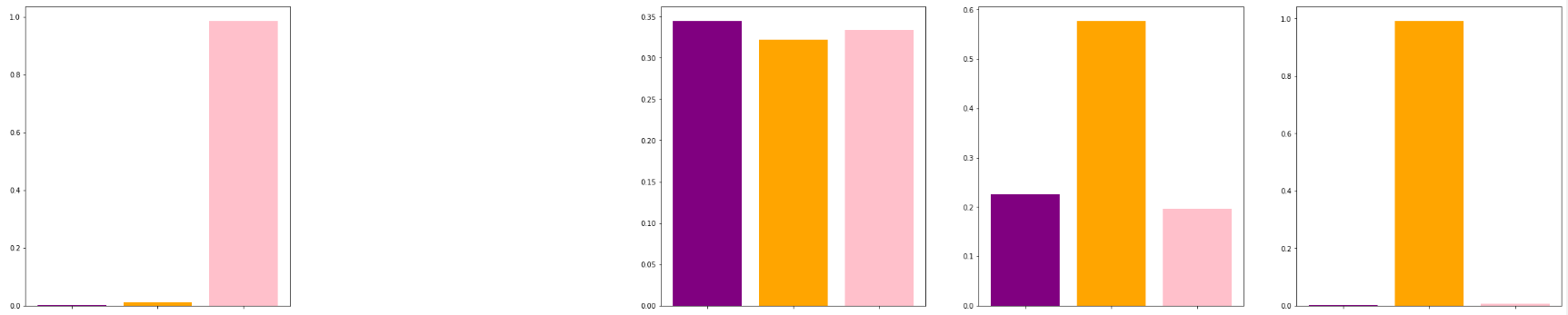}
    \caption{Pedagogical tutor policy for goal 1.}
  \end{subfigure}
  
  \centering
\caption{Pedagogical and naive tutor policies for goal 1. Left: action probability for the first pick. Right: action probability knowing the first pick is respectively purple, orange and pink.}
\label{fig:teacherpolicies}
\end{figure}

We first analyze the tutors' policies. In Fig.~\ref{fig:teacherpolicies}, we present the learned tutor policies for goal 1, which best illustrates the difference between a naive and a pedagogical tutor. Contrary to the naive tutor, the pedagogical tutor specifically avoids the demonstrations (orange, orange) because it does not allow to directly detect goal 1 from the first pick. Moreover, it avoids the ambiguous demonstration (orange, pink), which not only reaches goal 1 but also goal 2. This pedagogical tutor policy maximally avoids ambiguity in demonstrations.

\begin{figure}
\centering
\begin{subfigure}[b]{0.8\columnwidth}
    \includegraphics[width=\linewidth]{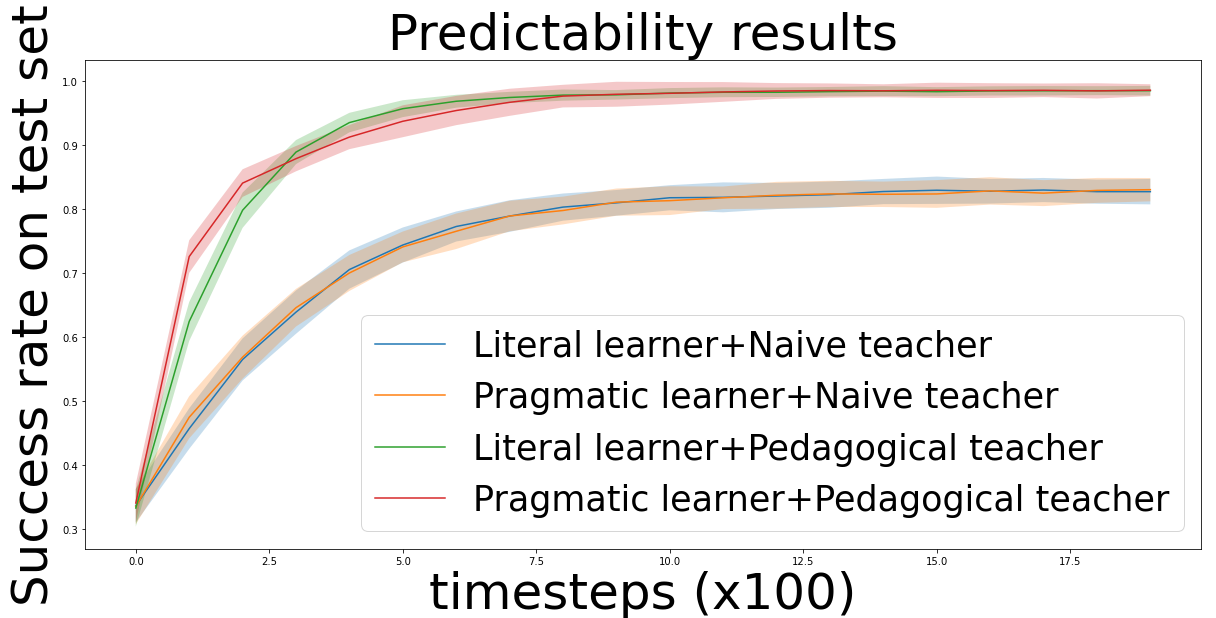}
    \caption{Predictability: can learners predict goals given demonstrations?}
  \end{subfigure}
  \hfill 
  \begin{subfigure}[b]{0.8\columnwidth}
    \includegraphics[width=\linewidth]{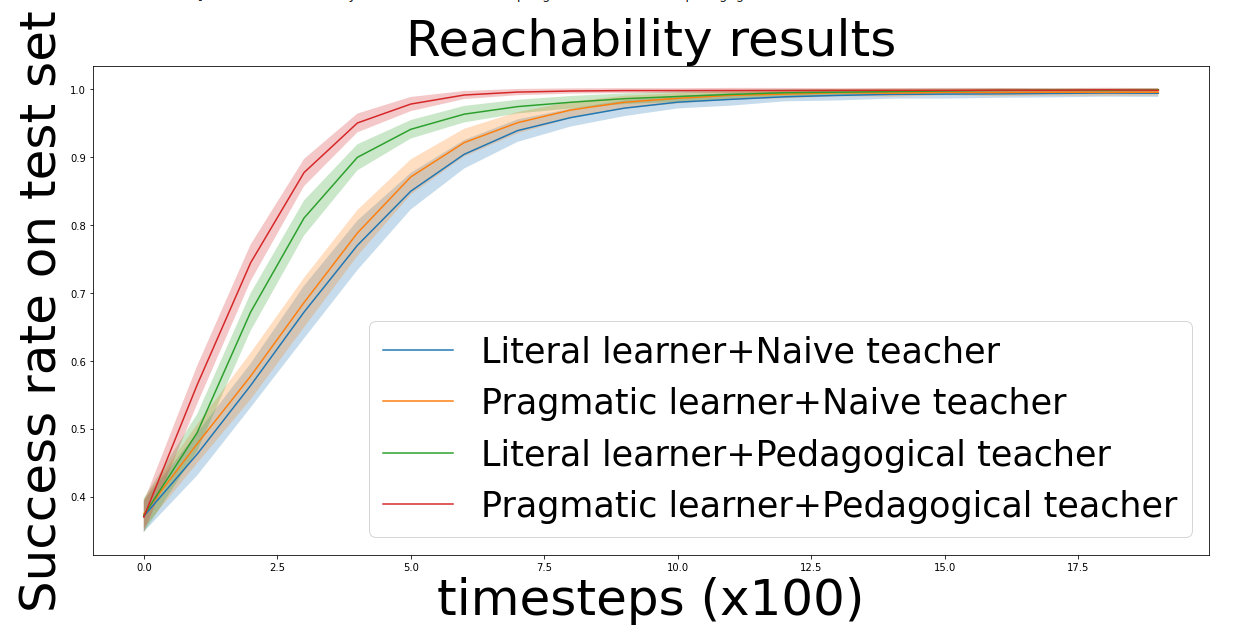}
    \caption{Reachability: can learners reach pursued goals?}
  \end{subfigure}
  
  \centering
\caption{Predictability (goal prediction accuracy) and reachability (goal reaching accuracy) results for all four possible combinations of experimental setups: literal/pragmatic learner and naive/pedagogical tutor.}
\label{fig:results}
\end{figure}

We then experiment with pedagogical/naive tutors and pragmatic/literal learners with the training loop presented in Sec.\ref{sec:policy}, and present our results on Fig.\ref{fig:results}. When a learner is trained with a pedagogical tutor, it consistently learns faster, and can predict the goals from all demonstrations, whereas it is not capable of disambiguating goal 1 from goal 2 with the (orange, pink) demonstration from a naive tutor. Moreover, a learner benefits from pragmatism if the tutor is pedagogical, resulting in the best tutor-learner combination.

\section{Conclusion and Future work}

In this short paper, we present preliminary results on implementing tutor pedagogy and learner pragmatism for a more efficient collaborative learning from demonstrations. Using a toy environment, we provide proof-of-concept for those pedagogical and pragmatic mechanisms. Future work will focus on implementing those mechanisms in more complex environments, using reinforcement learning as the core learning algorithm.

\section*{Acknowledgement}

This work was performed using HPC resources from GENCI-IDRIS (Grant 2022-A0131013011).

\bibliography{bibli}
\bibliographystyle{plain}

\end{document}